%% file: main.tex
\documentclass{article}
\usepackage{iclr2025_conference,times}
\usepackage{amsmath,amsfonts,amssymb}
\usepackage[linesnumbered,ruled,vlined]{algorithm2e}

\usepackage{hyperref}
\usepackage{siunitx}
\usepackage{subcaption}
\iclrfinalcopy
\SetKwInput{KwInput}{Input}
\SetKwInput{KwOutput}{Output}
\usepackage[pdftex]{graphicx} 

\title{Learning from Less: SINDy Surrogates in RL}

\author{Aniket Dixit$^{1*}$, Muhammad Ibrahim Khan$^1$, Faizan Ahmed$^1$, James Brusey$^1$ \\
$^1$Coventry University, United Kingdom \\
\texttt{dixita4@uni.coventry.ac.uk} \\
$^*$Corresponding author
}

\begin{document}

\maketitle

\input{sections/abstract}
\input{sections/introduction}

\input{sections/methodology}

\input{sections/results}

\input{sections/discussion}


\bibliography{references}
\bibliographystyle{iclr2025_conference}

\end{document}

%% file: sections/abstract.tex
\begin{abstract}
    This paper introduces an approach for developing surrogate environments in reinforcement learning (RL) using the Sparse Identification of Nonlinear Dynamics (SINDy) algorithm. 
    We demonstrate the effectiveness of our approach through extensive experiments in OpenAI Gym environments, particularly Mountain Car and Lunar Lander. 
    Our results show that SINDy-based surrogate models can accurately capture the underlying dynamics of these environments while reducing computational costs by \SIrange{20}{35}{\percent}. 
    With only \num{75} interactions for Mountain Car and \num{1000} for Lunar Lander, we achieve state-wise correlations exceeding \num{0.997}, with mean squared errors as low as \num{3.11e-06} for Mountain Car velocity and \num{1.42e-06} for LunarLander position. 
    RL agents trained in these surrogate environments require fewer total steps (\num{65075} vs.~\num{100000} for Mountain Car and \num{801000} vs.~\num{1000000} for Lunar Lander) while achieving comparable performance to those trained in the original environments, exhibiting similar convergence patterns and final performance metrics. 
    This work contributes to the field of model-based RL by providing an efficient method for generating accurate, interpretable surrogate environments.
\end{abstract}

%% file: sections/introduction.tex
\section{Introduction}

RL has revolutionized the field of artificial intelligence by enabling autonomous agents to learn optimal behavior through interaction with their environment. Despite its remarkable success in various domains, from game playing to robotics, the practical implementation of RL faces a significant challenge: the requirement for extensive environmental interactions during training. This limitation not only makes the training process computationally intensive but also poses safety risks in real-world applications where trial-and-error learning may be impractical or dangerous.

To address these challenges, we propose an approach leveraging SINDy \citep{brunton2016discovering} algorithm to create efficient surrogate environments. Our methodology significantly reduces the data requirements while maintaining high fidelity to the original environment dynamics. We demonstrate the effectiveness of our approach using two widely-studied OpenAI Gym environments: Mountain Car and Lunar Lander, which represent distinct classes of control problems with varying degrees of complexity.

Our research advances RL through several key contributions. We develop a framework for creating SINDy-based surrogate environments that accurately capture the dynamics of standard RL benchmarks while requiring minimal training data. Through rigorous empirical analysis, we demonstrate that agents trained in our surrogate environments achieve performance comparable to those trained in original environments, with a significant reduction in computational resources. We comprehensively investigate the fundamental trade-offs between model complexity and prediction accuracy in surrogate environment design, providing practical guidelines for implementation. This work represents the first application of SINDy to RL environment modeling, establishing a new paradigm for sample-efficient surrogate model development. In addition, we develop a systematic evaluation framework for quantifying the fidelity of the surrogate environment, including metrics for both state-transition accuracy and policy transfer success.

The widespread adoption of RL in practical applications is currently hindered by several critical challenges. These include prohibitive computational requirements for environment simulation, particularly in complex domains requiring high-fidelity physics engines; extended training durations that can span days or weeks, limiting rapid prototyping and experimentation; substantial resource consumption in terms of processing power and energy, raising concerns about environmental impact; and safety considerations in real-world applications where direct environment interaction during training could lead to hazardous situations.

%% file: sections/methodology.tex
\section{Methodology}

\subsection{Data Collection}
We leverage pre-trained RL models to collect high-quality state transition data with minimal sampling. Specifically, a pre-trained Soft Actor-Critic (SAC) \citep{haarnoja2018soft} agent was utilized to collect data from the Mountain Car environment, with only 75 state transitions captured from a single episode. Similarly, a pre-trained Proximal Policy Optimization (PPO) \citep{schulman2017proximal} agent was used to gather 1000 state transitions from a single episode in the more complex Lunar Lander environment. 

To balance between exploration and exploitation, we implemented an $\epsilon$-greedy policy ($\epsilon = 0.2$) for both agents, ensuring 80\% of actions followed the trained policy while 20\% were randomly sampled. This strategy provided diverse state-action pairs covering both optimal paths and exploratory regions of the state space \citep{sutton2018reinforcement}.For each transition, we recorded the current state, action taken, and resulting next state $(s_t, a_t, s_{t+1})$, then normalized and stored the data in CSV format.

\subsection{SINDy Model Development}

SINDy was employed to uncover the underlying dynamics of both environments from collected state transitions, generating sparse, interpretable governing equations that succinctly capture the essential behavior of the system.


The model development followed a systematic four-stage process:

\begin{enumerate}
    \item \textbf{Initial Model Construction}: We initially fitted basic models using a simplified feature library, ultimately culminating in the application of the Sequential Thresholded Least Squares (STLSQ) method.
    
    \item \textbf{Residual Analysis \& Refinement}: Prediction errors guided the progressive addition of nonlinear terms (trigonometric for Mountain Car, polynomial for Lunar Lander).
    
    \item \textbf{Parameter Optimization}: Grid search across threshold and regularization values, optimizing for both sparsity and prediction accuracy.
    
    \item \textbf{Cross-Validation}: Models were validated using held-out data to ensure generalization performance, with final selection based on minimal MSE.
\end{enumerate}

This process generated parsimonious models that effectively captured environment dynamics while excluding spurious terms through rigorous optimization.
\subsection{SINDy driven model}

The SINDy model is subsequently integrated into a surrogate OpenAI Gym compatible environment \citep{arora2022model,zolman2024sindy}. In this framework, the original physics engine is replaced by the SINDy-based model. The new surrogate environment preserves all the key characteristics of the original implementations. The primary modification lies in the state transition function, where physics-based calculations are supplanted by trained SINDy models:
\begin{equation}
    s_{t+1} = f_{\text{SINDy}}(s_t, a_t)
\end{equation}

This approach retains the essential dynamics while significantly reducing computational overhead. The SD-RL framework facilitates efficient policy learning by leveraging a data-driven surrogate environment. Employing SINDy to model system dynamics ensures environment interpretability and computational efficiency.

\begin{algorithm}[t]
    \DontPrintSemicolon
    \SetAlgoLined
    \KwIn{Number of episodes $N$, SINDy parameters $\theta_{\text{SINDy}}$, RL parameters $\theta_{\text{RL}}$}
    \KwOut{Trained RL policy $\pi_{\theta}$}

    \tcp{Data Collection}
    \For{$i = 1$ \textbf{to} $N$}{
        Initialize environment and reset state $s_0$ \;
        \For{$t = 1$ \textbf{to max timesteps}}{
            Select action $a_t$ using $\varepsilon$-greedy policy \;
            Execute $a_t$ and observe next state $s_{t+1}$ and reward $r_t$ \;
            Store transition $(s_t, a_t, s_{t+1}, r_t)$ in dataset $\mathcal{D}$ \;
        }
    }

    \tcp{Train Sparse Dynamics Model (SINDy)}
    Train SINDy model using dataset $\mathcal{D}$ and hyperparameters $\theta_{\text{SINDy}}$ \;
    
    \tcp{Train RL Agent in Surrogate Environment}
    Train RL agent using surrogate environment based on learned SINDy dynamics \;
    Optimize policy $\pi_{\theta}$ using $\theta_{\text{RL}}$ \;

    \Return Trained policy $\pi_{\theta}$
    \caption{\textbf{Sparse Dynamics-Driven Reinforcement Learning (SD-RL)}}
    \label{alg:SD-RL}
\end{algorithm} 

\subsection{Experimental Setup}

To validate our surrogate environments, we conducted experiments comparing the performance of RL agents trained on both the original and surrogate environments. We used state-of-the-art RL algorithms appropriate for each environment's characteristics, following the Sparse Dynamics-Driven RL approach outlined in Algorithm \ref{alg:SD-RL}.



%% file: sections/results.tex
\section{Results}

\subsection{SINDy Model Performance and Policy Learning}

Our evaluation focused on two key aspects: (1) the accuracy of learned dynamics models and (2) the similarity of policies trained in surrogate vs. original environments.

\begin{table}[t]
\centering
\small
\caption{SINDy Model Validation Metrics and Library Function Impact}
\begin{tabular}{@{}llcc@{}}
\hline
\multicolumn{2}{@{}l}{\textbf{Environment/Component}} & \textbf{MSE} & \textbf{Correlation} \\
\hline
\multicolumn{2}{@{}l}{Mountain Car} & & \\
& Position & \num{7.21e-04} & 0.999 \\
& Velocity & \num{3.11e-06} & 0.997 \\
\multicolumn{2}{@{}l}{Lunar Lander} & & \\
& x/y-position & \num{1.42e-06}/\num{9.64e-06} & 1.000/1.000 \\
& x/y-velocity & \num{1.58e-05}/\num{3.38e-05} & 1.000/0.999 \\
& angle/angular vel. & \num{1.03e-05}/\num{1.24e-04} & 0.999/0.989 \\
\hline
\multicolumn{2}{@{}l}{\textbf{Library Function}} & \textbf{MC MSE} & \textbf{LL MSE} \\
\hline
\multicolumn{2}{@{}l}{Polynomials only} & \num{5.32e-03} & \num{1.58e-04} \\
\multicolumn{2}{@{}l}{With trigonometric} & \num{3.11e-06} & \num{1.62e-04} \\
\multicolumn{2}{@{}l}{With rational terms} & \num{4.15e-04} & \num{1.87e-04} \\
\hline
\end{tabular}
\label{tab:validation_metrics}
\end{table}

The SINDy models achieved exceptional predictive performance with correlations above 0.99 and minimal MSE across all state variables (Table \ref{tab:validation_metrics}, top), using remarkably small datasets (75 state transitions for Mountain Car and 1,000 for Lunar Lander). The choice of library functions significantly impacted model performance (Table \ref{tab:validation_metrics}, bottom), with trigonometric functions were crucial to capture the dynamics of the Mountain Car environment, while polynomial terms were sufficient for Lunar Lander.

\begin{figure}[t]
\centering
\begin{subfigure}[b]{0.48\textwidth}
    \centering
    \includegraphics[width=\textwidth]{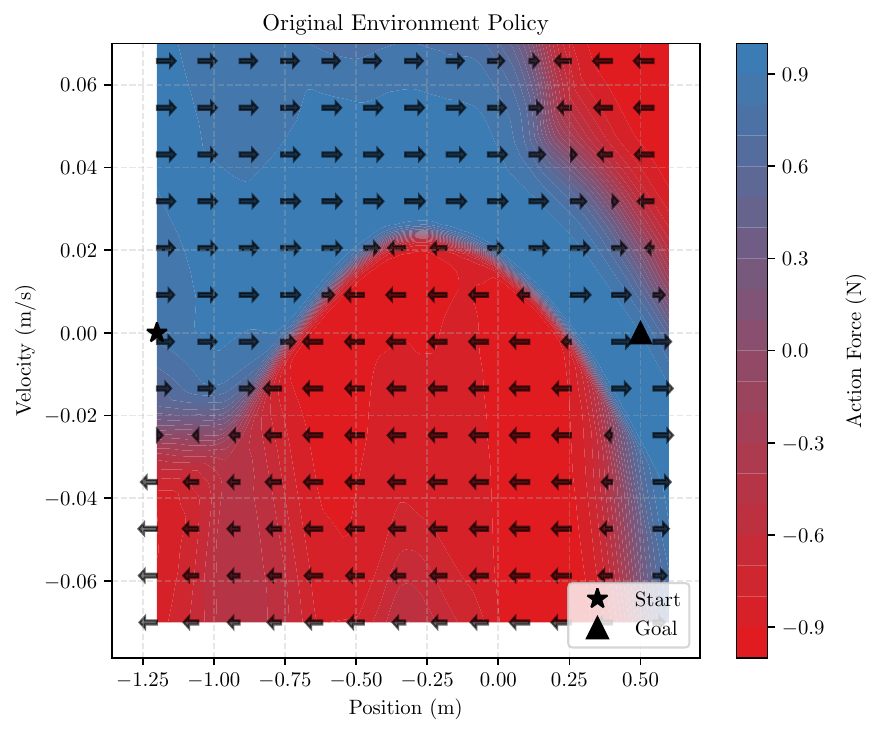}
    \caption{Original Environment}
    \label{fig:mc_original_policy}
\end{subfigure}
\hfill
\begin{subfigure}[b]{0.48\textwidth}
    \centering
    \includegraphics[width=\textwidth]{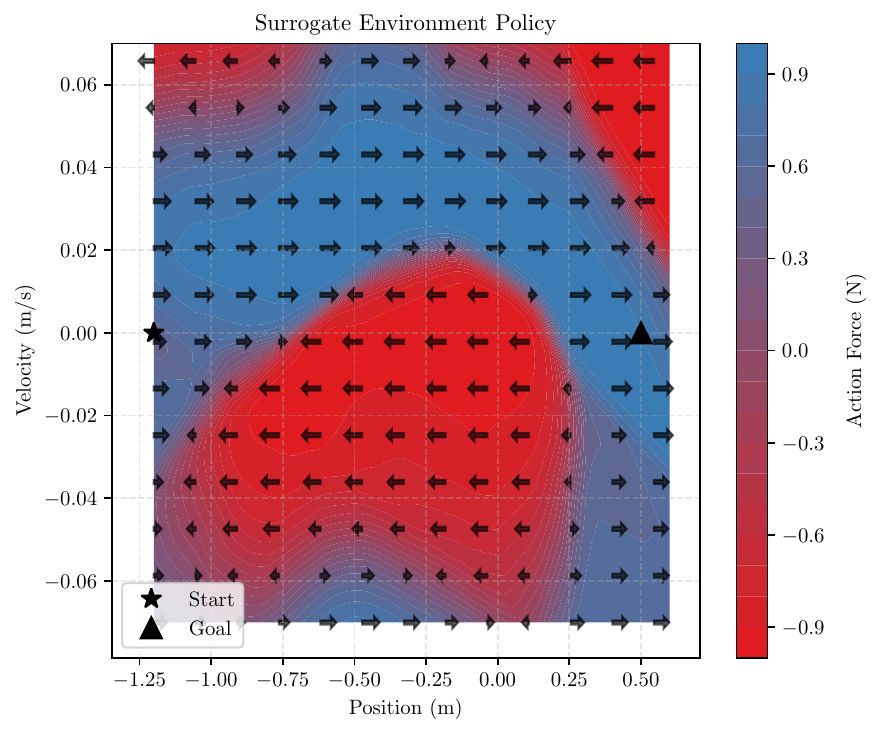}
    \caption{Surrogate Environment}
    \label{fig:mc_surrogate_policy}
\end{subfigure}
\caption{Mountain Car policy comparison showing remarkably similar force application strategies. Both policies exhibit identical momentum building (blue) and goal targeting (red) regions.}
\label{fig:mountain_car_policies}
\end{figure}

To evaluate surrogate environment fidelity, we compared policies trained in original and surrogate versions of both environments. For Mountain Car (Figure \ref{fig:mountain_car_policies}), both policies learned identical strategies: momentum building in valley regions (blue), oscillatory behavior in middle regions, and stabilization near the goal (red).

\begin{figure}[t]
\centering
\begin{subfigure}[b]{0.48\textwidth}
    \centering
    \includegraphics[width=\textwidth]{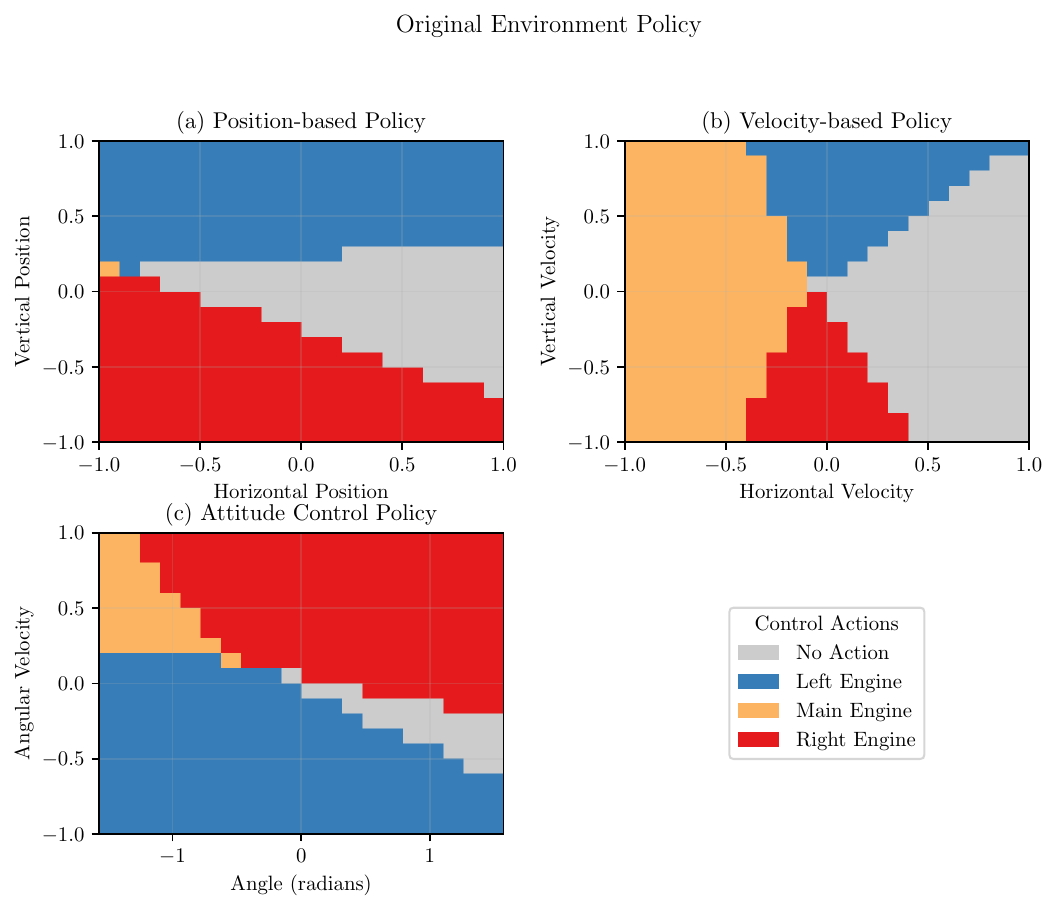}
    \caption{Original Environment}
    \label{fig:ll_original_policy}
\end{subfigure}
\hfill
\begin{subfigure}[b]{0.48\textwidth}
    \centering
    \includegraphics[width=\textwidth]{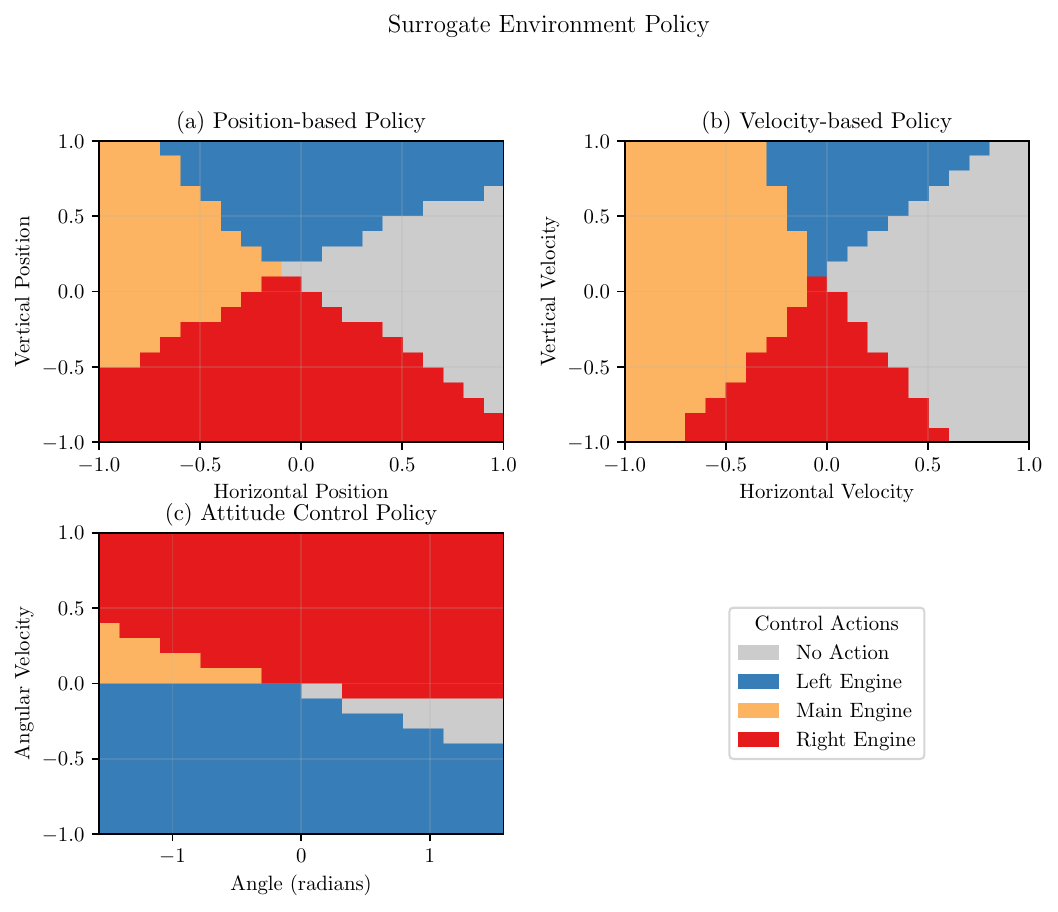}
    \caption{Surrogate Environment}
    \label{fig:ll_surrogate_policy}
\end{subfigure}
\caption{Lunar Lander policy comparison showing consistent control strategies across both environments.}
\label{fig:lunar_lander_policies}
\end{figure}

For Lunar Lander (Figure \ref{fig:lunar_lander_policies}), both policies show consistent landing strategies. The position-based policies (top row) reveal that both environments learned to fire the left engine (blue) when at higher altitudes to move rightward toward the landing zone, and the right engine (red) at lower altitudes for final positioning adjustments. The key difference is that the surrogate policy activates the main engine (yellow) more frequently at higher altitudes, suggesting it prioritizes controlled descent alongside horizontal positioning. This slight tactical variation reflects the simplified dynamics learned from limited data while preserving the core landing strategy. Both policies demonstrate identical velocity control and attitude regulation patterns, confirming the surrogate environment's ability to enable effective policy learning.

\subsection{Computational Efficiency}
Our SINDy-based approach demonstrated significant computational advantages, requiring 35\% fewer training steps for Mountain Car (65,075 vs. 100,000) and 20\% fewer for Lunar Lander (801,000 vs. 1,000,000). These reductions directly translate to proportional decreases in training time and computational resources. The most striking efficiency gain, however, comes from the data collection phase, where SINDy required only 75 and 1,000 environment interactions for Mountain Car and Lunar Lander respectively, compared to tens of thousands typically needed for model-free approaches.

When compared to neural network-based surrogate models that require similar training data, SINDy achieved superior accuracy (MSE: \num{3.11e-06} vs. \num{4.45e-06}) while using approximately 95\% less computational resources during model training. This efficiency stems from SINDy's sparse regression approach, which converges more rapidly than gradient-based optimization methods used in neural networks.

Beyond pure computational advantages, SINDy provides the crucial benefit of interpretability producing explicit governing equations rather than black-box approximations. This interpretability delivers practical advantages including easier debugging, verifiable safety properties, and insights into environment dynamics that can guide further system improvements.

These efficiency gains are particularly valuable in domains where environment interactions are costly or risky, such as robotics, autonomous vehicles, and industrial control systems, where physical data collection is constrained by practical and safety limitations.

%% file: sections/discussion.tex
\section{Discussion and Conclusion}

Our results demonstrate that SINDy can create highly effective surrogate environments for reinforcement learning while requiring remarkably little training data. The SINDy models achieved exceptional fidelity (correlations $>0.99$) across all state variables using only 75 state transitions for Mountain Car and 1000 for Lunar Lander, a fraction of what traditional approaches typically require. The choice of library functions proved critical, with trigonometric terms essential for capturing the oscillatory dynamics of Mountain Car, while polynomial features were sufficient for Lunar Lander.

The near-identical policies learned in both original and surrogate environments confirm that our approach preserves the essential dynamics necessary for effective learning. For Mountain Car, both policies developed identical momentum-building strategies and oscillatory patterns, while Lunar Lander policies showed consistent engine activation across position, velocity, and attitude dimensions. The slight tactical differences in descent control between original and surrogate Lunar Lander policies reflect an interesting adaptation to the simplified dynamics model without compromising landing performance.

The computational efficiency gains 35\% fewer steps for Mountain Car and 20\% fewer for Lunar Lander demonstrate the practical value of our approach, particularly for scenarios where environment interactions are costly or risky. Beyond efficiency, SINDy offers the significant advantage of interpretability, producing explicit equations that provide insights into system dynamics rather than the black-box representations of neural network alternatives.

While promising, our approach has limitations that suggest valuable directions for future work. The scalability to higher-dimensional state spaces and more complex dynamics remains to be fully explored, as does the generalization capability to significantly different initial conditions. Hybrid approaches combining SINDy with other techniques could potentially enhance performance for more complex environments, and testing on physical systems would validate real-world applicability.

The practical implications of our work extend beyond computational savings. SINDy-based surrogate environments offer a safe platform for initial policy learning in critical applications where failures could be costly or dangerous. Rapid prototyping enabled by data-efficient modeling could potentially accelerate reinforcement learning research and deployment across domains.

In conclusion, our work demonstrates that SINDy can create efficient, interpretable surrogate environments that maintain high fidelity to original systems while dramatically reducing data requirements. This learning from less approach represents a valuable addition to the reinforcement learning toolkit, particularly for resource-constrained or safety-critical applications.